\documentclass[journal]{IEEEtran}
\usepackage{graphicx}
\usepackage{cite}
\usepackage[T1]{fontenc}
\usepackage{amsmath,amssymb,amsfonts}
\usepackage{algorithmic}
\usepackage{textcomp}
\usepackage{booktabs}
\usepackage{xspace}
\usepackage{svg}
\usepackage{amsmath}
\usepackage{footnote}
\usepackage{makecell}
\usepackage{multirow}
\usepackage{enumitem}
\usepackage{textcomp}
\usepackage{array}
\usepackage[nolist,nohyperlinks]{acronym}
\usepackage{hyperref}
\usepackage{subcaption}
\makesavenoteenv{tabular}
\usepackage{doi}
\usepackage[acronym]{glossaries}

\acrodef{GI}{gastrointestinal}
\acrodef{AI}{Artificial Intelligence} 
\acrodef{CADx}{Computer Aided Diagnosis} 
\acrodef{ML}{Machine Learning}
\acrodef{DL}{Deep Learning}
\acrodef{CNN}{Convolutional Neural Network}
\acrodef{CRC}{Colorectal Cancer}
\acrodef{WCE}{Wireless Capsule Endoscopy}  
\acrodef{BoF}{Bag of Feature}  
\acrodef{GIANA}{Gastrointestinal Image ANAlysis} 
\acrodef{FCN}{Fully Convolutional Network}
\acrodef{ASPP}{Atrous Spatial Pyramidal Pooling}
\acrodef{SGDR}{Stochastic Gradient Descent with Restart}
\acrodef{AUC-ROC}{Area Under Curve - Receiver Operating Characteristic}
\acrodef{ROC}{Receiver Operating Curve}
\acrodef{SGD}{Stochastic Gradient Descent}
\acrodef{MSE}{Mean Square Error}
\acrodef{ReLU}{Rectified Linear Unit}
\acrodef{BN}{Batch Normalization}
\acrodef{ASPP}{Atrous Spatial Pyramid Pooling}
\acrodef{ReLU}{Rectified Linear Unit}
\acrodef{SOTA}{state-of-the-art}
\acrodef{FPS}{Frame Per Second}
\acrodef{FCM}{Fuzzy C-mean}
\acrodef{GAN}{Generative Adversarial Network}
\acrodef{mIoU}{mean Intersection over Union}
\acrodef{FCN}{Fully Convolutional Network}
\acrodef{DSC}{Dice Coefficient}
\acrodef{IoU}{Intersection over Union}

\def\BibTeX{{\rm B\kern-.05em{\sc i\kern-.025em b}\kern-.08em
    T\kern-.1667em\lower.7ex\hbox{E}\kern-.125emX}}
\begin{document}

\title{Real-Time Polyp Detection, Localization and Segmentation in Colonoscopy Using Deep Learning}

\author{Debesh Jha*,
Sharib Ali*,
Nikhil Kumar Tomar,
H{\aa}vard D. Johansen,
Dag Johansen,
Jens Rittscher,
Michael A. Riegler,
P{\aa}l Halvorsen

\thanks{Manuscript received 2021-02-02; accepted 2021-02-15; Date of Publication 2021-03-15.} 

\doi{10.1109/ACCESS.2021.3063716}

\thanks{D. Jha is with SimulaMet, Oslo, Norway and UiT The Arctic University of Norway, Troms{\o}, Norway.}
\thanks{S. Ali is with the Department of Engineering Science, University of Oxford, and Oxford NIHR Biomedical Research Centre, Oxford, UK.}
\thanks{N. K. Tomar is with SimulaMet, Oslo, Norway.}
\thanks{H.D. Johansen and D. Johansen are with UiT The Arctic University of Norway, Troms{\o}, Norway.}
\thanks{J. Rittscher is with the Department of Engineering Science, University of Oxford, Oxford, UK.}
\thanks{M. A. Riegler is with SimulaMet, Oslo, Norway.}
\thanks{P. Halvorsen is with SimulaMet and Oslo Metropolitan University, Oslo, Norway.}

\thanks{Corresponding authors: Debesh Jha (e-mail: debesh@simula.no), Sharib Ali (sharib.ali@eng.ox.ac.uk); * represents contributed equally to the manuscript}

}

\markboth{IEEE Access, Volume: 9,  Pages 40496 - 40510, Date of Publication: 04 March 2021, Electronic ISSN: 2169-3536, DOI: 10.1109/ACCESS.2021.3063716}
{Jha \MakeLowercase{\textit{et al.}}: Real-Time Polyp Detection, Localization and Segmentation in Colonoscopy Using Deep Learning}

\maketitle

\begin{abstract}
Computer-aided detection, localisation, and segmentation methods can help improve colonoscopy procedures. Even though many methods have been built to tackle automatic detection and segmentation of polyps, benchmarking of state-of-the-art methods still remains an open problem. This is due to the increasing number of researched computer vision methods that can be applied to polyp datasets. Benchmarking of novel methods can provide a direction to the development of automated polyp detection and segmentation tasks. Furthermore, it ensures that the produced results in the community are reproducible and provide a fair comparison of developed methods. 
In this paper, we benchmark several recent state-of-the-art methods using Kvasir-SEG, an open-access dataset of colonoscopy images for polyp detection, localisation, and segmentation evaluating both method accuracy and speed. Whilst, most methods in literature have competitive performance over accuracy, we show that the proposed ColonSegNet achieved a better trade-off between an average precision of 0.8000 and mean IoU of 0.8100, and the fastest speed of 180 frames per second for the detection and localisation task. Likewise, the proposed ColonSegNet achieved a competitive dice coefficient of 0.8206 and the best average speed of 182.38 frames per second for the segmentation task. Our comprehensive comparison with various state-of-the-art methods reveals the importance of benchmarking the deep learning methods for automated real-time polyp identification and delineations that can potentially transform current clinical practices and minimise miss-detection rates.

\end{abstract}

\IEEEpeerreviewmaketitle

\section{Introduction}
\label{sec:introduction}
\underline

\ac{CRC} has the third highest mortality rate among all cancers. The overall five-year survival rate of colon cancer is around 68\%, and stomach cancer is only around 44\%~\cite{asplund2018survival}. Searching for and removing precancerous anomalies is one of the best working methods to avoid \ac{CRC} based mortality. Among these abnormalities, polyps in the colon are important to detect because it can develop into the \ac{CRC} at late stage. Thus, an early detection of \ac{CRC} is crucial for survival. 
 
After modification in the lifestyle, the prevention from the \ac{CRC} is the screening of the colon regularly. Different research studies suggest that population-wide screening advances the prognosis and can even reduce the incidence of \ac{CRC}~\cite{holme2013flexible}. Colonoscopy is an invasive medical procedure where an endoscopist examines and operates on the colon using a flexible endoscope. It is considered to be the best diagnostic tool for colon examination for early detection and removal of polyps. Therefore, colonoscopic screening is the most preferred technique among gastroenterologists.

Polyps are abnormal growths of tissue protruding from the mucous membrane. They can occur anywhere in the \ac{GI} tract but are mostly found in the colorectal area and are often considered a predecessor of \ac{CRC}~\cite{jha2019resunet++,holzheimer2001surgical}. Polyps may be pedunculated (having a well-defined stalk) or sessile (without a defined stalk). The colorectal polyps can be categorised into two classes: non-neoplastic and neoplastic. Non-neoplastic polyps are further sub-categorised into hyperplastic, inflammatory, and hamartomatous polyps. These types of polyps are non-cancerous and not harmful. Neoplastic is further sub-categorised into adenomas and serrated polyps. These polyps can develop into the risk of cancer. Based on their size, colorectal polyps can be categorised into three classes, namely, diminutive ($\leq$5mm), small (6 to 9 mm), and advanced (large) ($\geq$10mm)~\cite{lee2016resection}. Usually, larger polyps can be detected and resected.

There exists a significant risk with small and diminutive colorectal polyps~\cite{ponugoti2017risk}. A polypectomy is a technique for the removal of small and diminutive polyps. There are five different polypectomy techniques for resection of diminutive polyps, namely, cold forceps polypectomy, hot forceps polypectomy, cold snare polypectomy, hot snare polypectomy, and endoscopic mucosal resection~\cite{lee2016resection}. Among these techniques, cold snare polypectomy is considered best polypectomy technique for resectioning small colorectal polyps~\cite{tranquillini2018best}.

Colonoscopy is an invasive procedure that requires high-quality bowel preparation as well as air insufflation during examination~\cite{kronborg2007population}. It is both an expensive and time-demanding procedure.  Nevertheless, on average, 20\% of polyps are missed during examinations. The risk of getting cancer therefore relates to the individual endoscopists' ability to detect polyps~\cite{kaminski2010quality}. Recent studies have shown that new endoscopic devices and diagnostic tools have improved the adenoma detection rate and polyp detection rate~\cite{castaneda2018new,matyja2019improve}. However, the problem of over-looked polyps remains the same. 


The colonoscopy videos recorded at the clinical centers store a significant amount of colonoscopy data. However, the collected data are not used efficiently as they are labour intense for the endoscopists~\cite{riegler2017eir}. Thus, a second review of videos are often not done. This might lead to missed detection at an early stage largely. 
Automated data curation and annotation of video data is a prerequisite for building reliable \ac{CADx} systems that can help to assess clinical endoscopy more thoroughly~\cite{de2018methodology}. A fraction of the collected colonoscopy data can be curated to develop computer-aided systems for automated detection and delineation of polyps either during the clinical procedure or after the reporting. At the same time, to build a robust system, it is vital to incorporate data variability related to patients, endoscopic procedure, and endoscope manufacturers. Even though recent developments in computer vision and system designs have enabled us to built accurate and efficient systems, these largely depend on the data availability as most recent methods are data voracious. The lack of availability of public datasets~\cite{shin2018abnormal} is a critical bottleneck to accelerate algorithm development in this realm.

In general, curating medical datasets are challenging and it requires domain knowledge expertise. Reaching a consensus to achieve ground truth labels from different experts on the same dataset is again another obstacle. Typically, in colonoscopy, smaller polyps or flat/sessile polyps that are usually missed out during a procedure can be difficult to observe even during manual labeling. Other challenges include the patient variability and presence of different sizes, shapes, textures, colors, and orientations of these polyps~\cite{jha2019resunet++}. Therefore, during polyp data curation and developing of automated systems for the colonoscopy, it is vital that all various challenges often come along routine colonoscopy has to be taken into consideration.

Automatic polyp detection and segmentation systems based on \ac{DL} have a high overall performance in both colonoscopy images and colonoscopy videos~\cite{lee2020real,wang2018development}. Ideally, the automatic \ac{CADx} systems for polyps detection, localisation, and segmentation should have: 1) consistent performance and improved robustness to patient variability, i.e., the system should be able to produce reliable outputs, 2) high overall performance surpassing the set bar for algorithms, 3) real-time performance required for clinical applicability, and 4) easy-to-use system that can provide with clinically interpretable outputs. Scaling this to a population sized cohort is also a very resource-demanding and incurs enormous costs. As a first step, we therefore target the detection, localisation, and segmentation of colorectal polyps known as precursors of \ac{CRC}. The reason for starting with this scenario is that most colon cancers arise from benign adenomatous polyps (around 20\%) containing dysplastic cells. Detection and removal of polyps prevent the development of cancer, and the risk of getting \ac{CRC} in the following 60 months after a colonoscopy depends largely on the endoscopist ability to detect polyps~\cite{kaminski2010quality}. 

Detection and localisation of polyps are usually critical during routine surveillance and to measure the polyp load of the patient at the end of the surveillance while pixel-wise segmentation becomes vital to automate the polyp boundary delineation during the surgical procedures or radio-frequency ablations. In this paper, we evaluate \ac{DL} methods for both detection (and localisation referring to bounding box detection) and segmentation (pixel-wise classification or semantic segmentation) SOTA methods on Kvasir-SEG dataset~\cite{jha2020kvasir} to provide a comprehensive benchmark for the colonoscopy images. The main aim of the paper is to establish a new strong benchmark with existing successful computer vision approaches. 
Our contributions can be summarised as follows:
\begin{itemize}
     
    \item We propose ColonSegNet, an encoder-decoder architecture for segmentation of colonoscopic images. The architecture is very efficient in terms of processing speed (i.e., produces segmentation of colonoscopic polyp in real-time) and competitive in terms of performance.
     
   \item A comprehensive comparison of the state-of-the-art computer vision baseline methods on the Kvasir-SEG dataset is presented. The best approaches show real-time performance for polyp detection, localisation, and segmentation.  
    
    
    \item We have established strong benchmark for detection and localisation on the Kvasir-SEG dataset. Additionally, we have extended segmentation baseline as compared to~\cite{jha2019resunet++,jha2020kvasir,jha2021comprehensive}. These benchmarks can be useful to develop reliable and clinically applicable methods. 
    
    \item Detection, localisation, and semantic segmentation performances are evaluated on standard computer vision metrics.
    
    \item {Detailed analysis have been presented with the specific focus on the best and worst performing cases that will allow to dissect method success and failure modes required to accelerate algorithm development.} 
\end{itemize}

The rest of the paper is organized as follows: In Section~\ref{sec:relatedwork}, we present related work in the field. In Section~\ref{sec:materials}, we present the material. Section~\ref{method} presents both detection, localisation, and segmentation methods. Result are presented in Section~\ref{sec:results}. Discussion on the best performing detection, localisation, and semantic segmentation approaches are presented in Section~\ref{sec:discussion} and finally a conclusion is provided in the Section~\ref{sec:conclusion}. 


\section{Related work}
\label{sec:relatedwork}
\begin{table*}[t!]
\footnotesize
\centering
 \caption{Available endoscopic datasets}
    \label{tab:datasetsummary}
  \def\arraystretch{1.2}
  \begin{tabular}{l|c|c|c|c|c}
\toprule
\textbf{Dataset} & \textbf{Organ} & \textbf{Source} & \textbf{Findings} & \textbf{Dataset content} &  \textbf{Task type} \\ \midrule

Kvasir-SEG~\cite{jha2020kvasir} & Large bowel & WL$^\diamond$ & Polyp & \begin{tabular}[c]{@{}l@{}}1000 images \end{tabular} & \begin{tabular}[c]{@{}l@{}}  Detection, localisation\\ \& segmentation \end{tabular} \\ \hline 

Kvasir~\cite{pogorelov2017kvasir} & Whole GI & WL$^\diamond$& \begin{tabular}[c]{@{}l@{}}Polyps, esophagitis, ulcerative colitis,\\ z-line, pylorus, cecum, dyed polyp,\\ dyed resection margins, stool \end{tabular} & 8,000 images & Classification \\ \hline 

Nerthus~\cite{pogorelov2017nerthus} & Large bowel & WL$^\diamond$ & \begin{tabular}[c]{@{}l@{}}Stool - categorization of bowel cleanliness \end{tabular} & 21 videos & Classification \\ \hline 


HyperKvasir~\cite{borgli2019hyper} & Whole GI & WL$^\diamond$ & \begin{tabular}[c]{@{}l@{}}16 different classes from upper GI \& 24\\ different classes from lower GI tract\end{tabular} & \begin{tabular}[c]{@{}l@{}}110,079 images\\ \& 373 videos \end{tabular} & Classification \\ \hline 

ETIS-Larib~\cite{silva2014toward} & Colonoscopy & WL$^\diamond$ & Polyp & 196 images & Segmentation \\ \hline

CVC-Clinic~\cite{bernal2015wm} & Colonoscopy & WL$^\diamond$& Polyp &\begin{tabular}[c]{@{}l@{}} 612 images \end{tabular} & Segmentation \\\hline 

KvasirCapsule~\cite{smedsrud2020kvasir} & Whole GI & VCE & \begin{tabular}[c]{@{}l@{}}13 different classes of GI anomalies\end{tabular} &\begin{tabular}[c]{@{}l@{}} 4,820,739 images\\ \& 118 videos \end{tabular} & Classification \\ \hline

EDD 2020~\cite{ali2020endoscopy} & \begin{tabular}[c]{@{}l@{}}Entire GI \end{tabular} & \begin{tabular}[c]{@{}l@{}} NBI$^\dag$,\\ WL$^\diamond$ \end{tabular} &\begin{tabular}[c]{@{}l@{}} Polyp, Barrett's esophagus, high-grade\\  dysplasia, suspicious (low-grade), cancer \end{tabular}& $386$ images & \begin{tabular}[c]{@{}l@{}} Detection, localisation\\ \& segmentation \end{tabular}\\ \hline

Kvasir-Instrument~\cite{jha2021kvasir} &  Large Bowel &  WL$^\diamond$ & Tools and instruments &590 images  &\begin{tabular}[c]{@{}l@{}} Detection, localization,\\ Segmentation\end{tabular}\\

\bottomrule
\multicolumn{2}{l}{$^\dag$ Narrow band imaging \hspace{.1cm}
	   $^\diamond$White light imaging}\\
\end{tabular}
\end{table*}

Automated polyp detection has been an active topic for research 
over the last two decades and considerable work has been done to develop efficient methods and algorithms. Earlier works were especially focused on polyp color and texture, using handcrafted descriptors-based feature learning~\cite{karkanis2003computer,ameling2009texture}. 
More recently, methods based on \acp{CNN} have received significant attention~\cite{tajbakhsh2016convolutional,shin2016deep}, and have
been the go to approach for those competing in  public challenges~\cite{bernal2017comparative,ali2020objective}.

Wang et al.~\cite{wang2015polyp} designed algorithms and developed software modules for fast polyp edge detection and polyp shot detection, including a polyp alert software system. Shin et al.~\cite{shin2018automatic} have used region-based \ac{CNN} for automatic polyp detection in colonoscopy videos and images. They used Inception ResNet as a transfer learning approach and post-processing techniques for reliable polyp detection in colonoscopy. Later on, Shin et al.~\cite{shin2018abnormal} used generative adversarial network~\cite{goodfellow2014generative}, where they showed that the generated polyp images are not qualitatively realistic; however, they can help to improve the detection performance. Lee et al.~\cite{lee2020real} used YOLO-v2~\cite{redmon2016you,redmon2017yolo9000} for the development of polyp detection and localisation algorithm. The algorithm produced high sensitivity and near real-time performance. Yamada et al.~\cite{yamada2019development} developed an artificial intelligence system that can automatically detect the sign of \ac{CRC} during colonoscopy with high sensitivity and specificity. They claimed that their system could aid endoscopists in real-time detection to avoid abnormalities and enable early disease detection. 



In addition to the work related to automatic detection and localisation, pixel-wise classification (segmentation) of the disease provides an exact polyp boundary and hence is also of high significance for clinical surveillance and procedures. 
Bernel et al.~\cite{bernal2017comparative} presented the results of the automatic polyp detection subchallenge, which was the part of the endoscopic vision challenge at the Medical Image Computing and Computer Assisted Intervention (MICCAI) 2015 conference. This work compared the performance of eight teams and provided an analysis of various detection methods applied on the provided polyp challenge data. Wang et al.~\cite{wang2018development} proposed a \ac{DL}-based SegNet~\cite{badrinarayanan2017segnet} that had a real-time performance with an inference of more than 25 frames per second. Geo et al.~\cite{guo2019giana} used fully convolution dilation networks on the Gastrointestinal Image ANAlysis (GIANA) polyp segmentation dataset. Jha et al.~\cite{jha2019resunet++} proposed ResUNet++ demonstrating 10\% improvement compared to the widely used UNet baseline on Kvasir-SEG dataset. They also further applied the trained model on the CVC-ClinicDB~\cite{bernal2015wm} dataset showing more than 15\% improvement over UNet. Ali et al.~\cite{ali2020objective} did a comprehensive evaluation for both detection and segmentation approaches for the artifacts present clinical endoscopy including colonoscopy data~\cite{ali2019endoscopy}. Wang et al.~\cite{wang2020boundary} proposed a boundary-aware neural network (BA-Net) for medical image segmentation. BA-Net is an encoder-decoder network that is capable of capturing the high-level context and preserving the spatial information. Later on, Jha et al.~\cite{jha2020doubleu} proposed DoubleUNet for the segmentation, which was applied to four biomedical imaging datasets. The proposed DoubleUNet is the combination of two UNet stacked on top of each other with some additional blocks. Experimental results on CVC-Clinic and ETIS-Larib polyp datasets show the \ac{SOTA} performances. In addition to the related work on polyp segmentation, there are studies on segmentation approaches~\cite{minaee2020image,baldeon2020adaresu,saeedizadeh2020covid,meng2020cnn}.

Datasets has been instrumental for medical research.  Table~\ref{tab:datasetsummary} shows the list of the available endoscopic image and video datasets. Kvasir-SEG, ETIS-Larib, and CVC-ClinicDB contain colonoscopy images, whereas Kvasir, Nerthus, and HyperKvasir contain the images from the whole \ac{GI}. KvasirCapusle contains images from video capsule endoscopy. All the dataset contains images acquired from conventional White Light (WL) imaging technique except the EDD dataset, where it contains images from both WL imaging and Narrow Band Imaging (NBI) techniques. All of these datasets contain at least a polyp class. Out of nine available datasets, Kvasir-SEG~\cite{jha2020kvasir}, ETIS-Larib~\cite{silva2014toward}, and CVC-ClinicDB~\cite{bernal2015wm} has manually labeled ground truth masks. Among them, Kvasir-SEG offers the most number of annotated samples providing both ground truth masks and bounding boxes offering detection, localisation, and segmentation task. All of the datasets are publicly available. 

Dataset development, benchmarking of the methods, and evaluation are critical in the medical imaging domain. It inspires the community to build clinically transferable methods on a well-curated and standardised dataset. Due to the lack of benchmark papers, it becomes utmost difficult to understand the clear strength of methods in the literature. New algorithm developments demonstrating its translational abilities in clinics is thus very minimal. Data science challenges do offer some insight, however, a comprehensive analysis on various different aspects such as detection, localisation, segmentation, and inference time estimation are still not covered by the most.

Inspired by the previous benchmark for polyp detection~\cite{bernal2017comparative}, endoscopic artifact detection ~\cite{ali2019endoscopy}, endoscopic disease detection and segmentation ~\cite{ali2020endoscopy}, endoluminal scene object segmentation~\cite{vazquez2017benchmark}, and endoscopic instrument segmentation~\cite{ross2020comparative}, we introduce a new benchmark for the automatic polyp detection, localisation and segmentation using publicly available Kvasir-SEG dataset.



\section{Materials -- Dataset}
\label{sec:materials}
\begin{figure}[t!]
    \centering
    \includegraphics[trim=0cm 0cm 0cm 0cm,clip=true,scale=0.24]{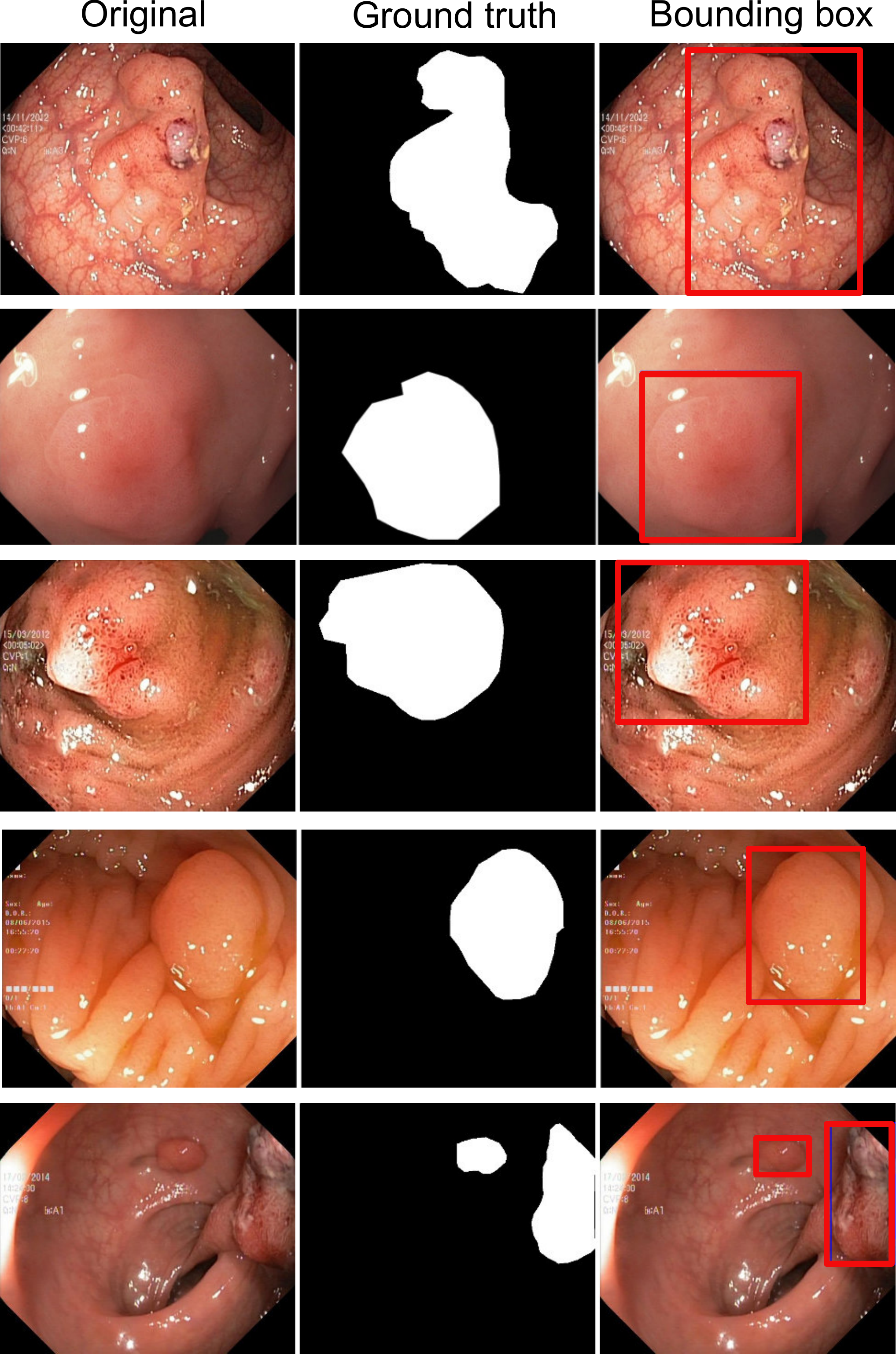}
    \caption{{Sample images from Kvasir-SEG dataset:} Annotated masks (2nd column) and bounding boxes (3rd column) for selected samples.}
    \label{fig:kvasir_sample}
\end{figure}



We have used the Kvasir-SEG~\cite{jha2020kvasir} for detection, localisation, and segmentation tasks. Figure~\ref{fig:kvasir_sample} shows the image, ground truth information, and their detection (their localised bounding boxes in red). This dataset is the outcome of an initiative for open and reproducible results. It contains 1000 polyp images acquired by high-resolution electromagnetic imaging system, i.e., ScopeGuide, Olympus Europe, their corresponding masks and bounding box information. The images and their ground truths can be used for the segmentation task, whereas the bounding box information provides an opportunity for the detection task. The resolution of the images in this dataset ranges from  $332\times 487$ to $1920\times 1072$ pixels. The dataset can be downloaded at \url{https://datasets.simula.no/kvasir-seg/}. The dataset includes images of 700 large polyps ($> 160 \times 160$ pixels), 323 medium sized polyps (> $64\times 64$ pixels and $\leq 160 \times 160$ pixels) and 48 small polyps ($\leq 64\times 64$ pixels). In total, the dataset consists of 1072 images of polyps with segmentation masks and bounding boxes.

\section{Method}

\label{method}
\begin{figure}
    \centering
    \includegraphics[scale=0.13]{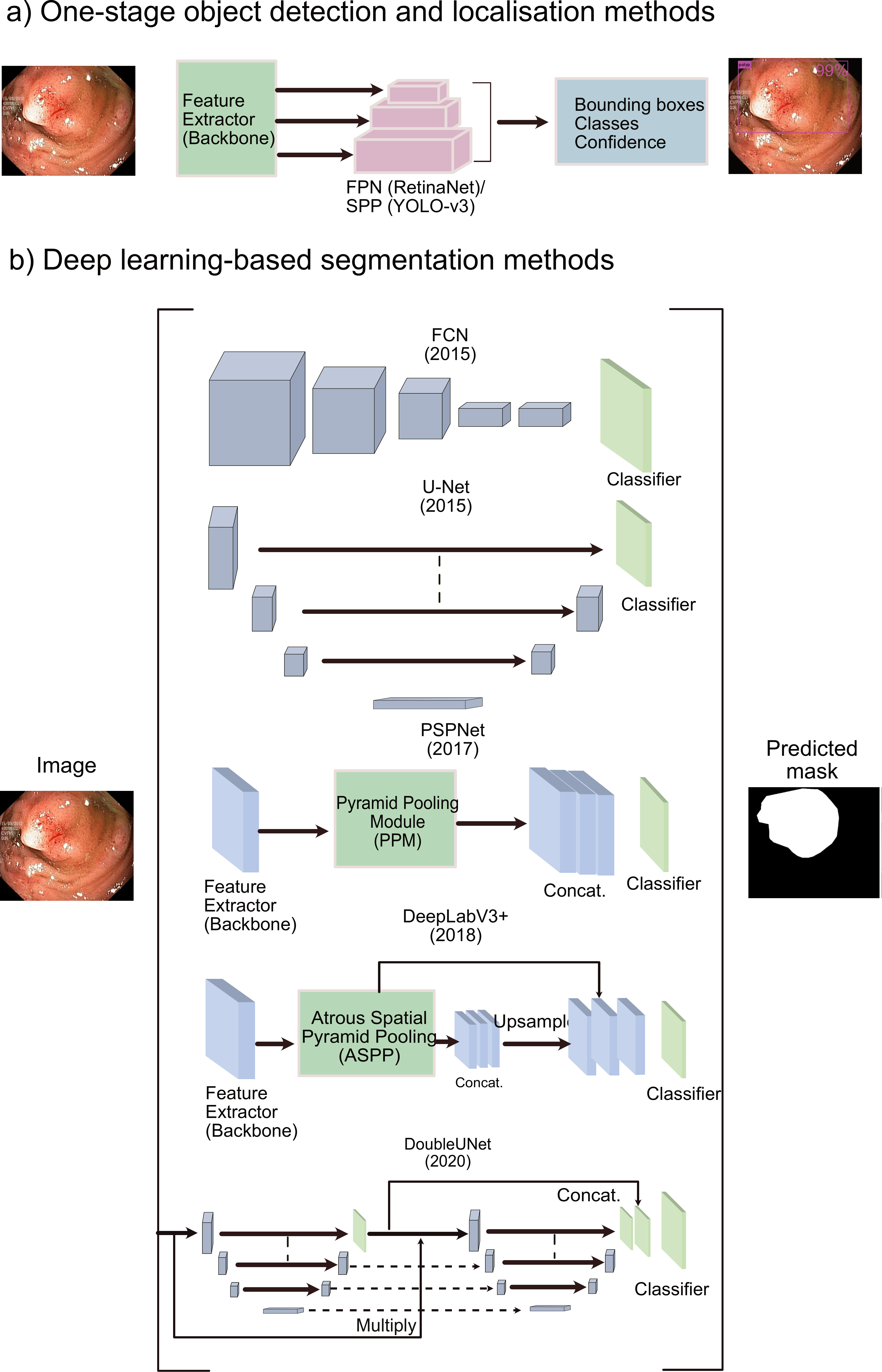}
    \caption{Baseline detection, localisation and semantic segmentation method summary.}
    \label{fig:baseline}
\end{figure}

Detection methods aim to predict the object class and regress bounding boxes for localisation, while segmentation methods aim to classify the object class for each pixel in an image. In Figure~\ref{fig:kvasir_sample}, ground truth masks for segmentation task are shown in 2nd column while corresponding bounding boxes for the detection task are in 3rd column.  This section describes the baseline methods for detection, localisation and segmentation methods used for the automated detection and segmentation of polyp in the Kvasir-SEG dataset.

\subsection{Detection and localisation baseline methods}
Detection methods consist of input, backbone, neck, and head. The input can be images, patches, or image pyramids. The backbone can be different \ac{CNN} architectures such as VGG16, ResNet50, ResNext-101, and Darknet. The neck is the subset of the backbone network, which could consist of FPN, PANet, and Bi-FPN. The head is used to handle the prediction boxes that can be one stage detector for dense prediction (e.g., YOLO, RPN, and RetinaNet~\cite{lin2017focal}), and two-stage detector with the sparse prediction (e.g., Faster R-CNN~\cite{ren2015faster}  and RFCN~\cite{dai2016r}).
Recently, one stage methods have attracted much attention due to their speed and ability to obtain optima accuracy. This has been possible because recent networks utilise feature pyramid networks or spatial-pyramid pooling layers to predict candidate bounding boxes which are regressed by optimising loss functions (see Figure~\ref{fig:baseline}).

In this paper, we use EfficientDet~\cite{tan2020efficientdet} which uses EfficientNet~\cite{tan2019efficientnet}, as the backbone architecture, bi-directional feature  pyramid network (BiFPN) as the feature network, and shared class/box prediction network. Additionally, we also use Faster R-CNN~\cite{ren2015faster}, which uses region proposal network (RPN), as the proposal network and Fast R-CNN~\cite{girshick2015fast} as the detector network.  
Moreover, we use YOLOv3~\cite{redmon2018yolov3} that utilises multi-class logistic loss (\textit{binary cross-entropy} for classification loss and \textit{mean square error} for regression loss) modeled with regularizers such as objectness prediction scores. 
Furthermore, we also used YOLOv4~\cite{bochkovskiy2020yolov4}, which  utilises an additional bounding box regressor based on the \ac{IoU} and a cross-stage partial connections in their backbone architecture. Additionally, YOLOv4 allows on fly data augmentation, such as mosaic and cut-mix.

RetinaNet~\cite{lin2017focal} takes into account the data driven property that allows the network to focus on ``hard'' samples for improved accuracy. The easy to adapt backbones for feature extraction at the beginning of the network provides the opportunity to experiment with deeper and varied architectures such as ResNet50, and ResNet101 for RetinaNet and 53 layered Darknet53 backbone for YOLOv3 and YOLOv4 architecture. To tackle the different aspect ratio problem, for both one stage networks, optimal anchor boxes~\cite{ren2015faster} are searched and pre-defined for the provided data to tackle large variance of scale and aspect ration of boxes.  Table~\ref{table:parametersuseddetection} shows the hyperparameter used by each of the object detection methods for the detection task. 

\subsection{Segmentation baseline methods}
In the past years, data-driven approaches using \acp{CNN} have changed the paradigm of computer vision methods, including segmentation. An input image can be directly be fed to convolution layers to obtain feature maps, which can be later upsampled to predict pixel-wise classification providing object segmentation. Such networks learn from available ground truth labels and can be used to predict labels from other similar data. A \ac{FCN} based segmentation was first proposed by  Long et al.~\cite{long2015fully} that can be trained end-to-end. Ronneberger et al.~\cite{ronneberger2015u} modified and extended the FCN architecture to a UNet architecture. The UNet consist of an analysis (\textit{encoder}) and a synthesis (\textit{decoder}) path. In the analysis path of the network, deep features are learnt, whereas in the synthesis path segmentation is performed on the basis of the learnt features.

Pyramid Scene Parsing Network (PSPNet)~\cite{zhao2017pyramid} introduced a pyramid pooling module aimed at aggregating global context information from different regions which are upsampled and concatenated to form the final feature representation. A final per-pixel prediction is obtained after a convolution layer (see Figure~\ref{fig:baseline}, third architecture). For feature extraction, we have used the ResNet50 architecture pretrained on imageNet. Similar to the UNet architecture, DeepLabV3+~\cite{chen2018encoder} is an encoder-decoder network. However, it utilizes atrous separable convolutions and spatial pyramid pooling (see Figure \ref{fig:baseline}, last architecture) for fast inference and improved accuracy. Atrous convolution controls the resolution of features computed and adjust the receptive field to effectively capture multi-scale information. In this paper, we have used an output stride of 16 for both encoder and decoder networks of DeepLabV3+ and have experimented on both ResNet50 and ResNet101 backbones.

ResUNet~\cite{zhang2018road} integrates the power of both UNet and residual neural network. ResUNet++~\cite{jha2019resunet++} is the improved version of ResUNet architecture. It has additional layers including squeeze-and-excite block, \ac{ASPP}, and attention block. These additional layers helps learning the deep features that are capable of improved prediction of pixels for object segmentation tasks. DoubleU-Net~\cite{jha2020doubleu} consists of two modified UNet architecture. It uses VGG-19 pretrained on ImageNet~\cite{deng2009imagenet} as the first encoder. The main reason behind using VGG-19 (similar to UNet~\cite{simonyan2014very}) was that it is a lightweight model. The additional component in the DoubleUNet are squeeze-and-excite block, and \ac{ASPP} block. High-Resolution Network (HRNet)~\cite{wang2020deep} maintains high-resolution representation convolution in parallel and interchange the information across the resolution continuously. This is one of the most recent and popular method in the literature. Furthermore, we have used UNet with ResNet34 as a backbone network and trained the model to compare with the other state-of-the-art semantic segmentation networks.

Table~\ref{table:parametersusedsegmentation} shows the hyperparameters used for each of the semantic segmentation based benchmark methods used. From the table, we can see that number of trainable parameters of the baseline methods are large. A high number of trainable parameters in the network makes it complex, leading to a lower frame rate. It is therefore essential to design an efficient, lightweight architecture that can provide a higher frame rate and better performance. In this regard, we propose a novel architecture, ColonSegNet, that requires only few number of training parameters, which can save training and inference time. More details about the architecture can be found in the below section. 

\begin{figure}[t!]
    \centering
    \includegraphics[width=0.9\linewidth]{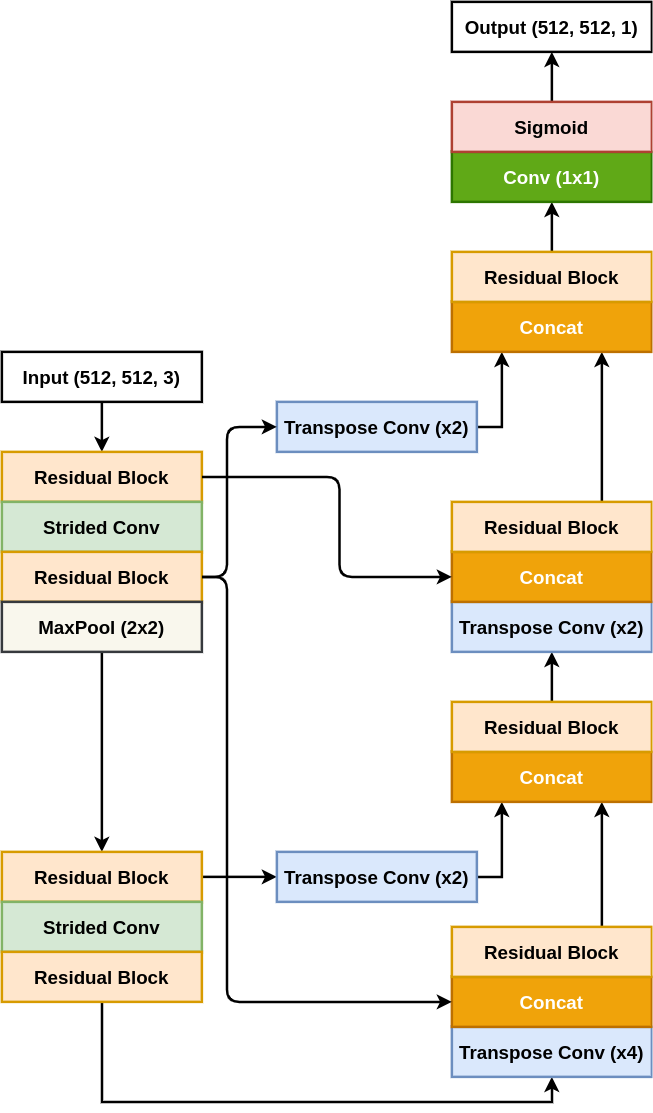}
    \caption{Block diagram of ColonSegNet}
    \label{fig:colonsegnet_diagram}
\end{figure}

\subsection{ColonSegNet}
Figure~\ref{fig:colonsegnet_diagram} shows the block diagram of the proposed ColonSegNet. It is an encoder-decoder that uses residual block~\cite{he2016deep} with squeeze and excitation network~\cite{hu2018squeeze} as the main component. The network is designed to have very few trainable parameters as compared to other networks baseline networks such as U-Net~\cite{ronneberger2015u}, PSPNet~\cite{zhao2017pyramid}, DeepLabV3+~\cite{chen2018encoder}, and others.  The use of fewer trainable parameters makes the proposed architecture a very light-weight network that leads to real-time performance.

The network consists of two encoder blocks and two decoder blocks. The encoder network learns to extract all the necessary information from the input image, which is then passed to the decoder. Each decoder block consists of two skip connections from the encoder. The first is a simple concatenation, and the second skip connection passed through a transpose convolution to incorporates multi-scale features in the decoder. These multi-scale features help the decoder to generate more semantic and meaningful information in the form of a segmentation mask.

The input image is fed to the first encoder, which consists of two residual blocks and a $3 \times 3$ strided convolution in between them.  This layer is followed by a $2 \times 2$ max-pooling. Here, the output feature map spatial dimensions are reduced to $\frac{1}{4}$ of the input image.  The second encoder consists of two residual blocks and a $3 \times 3$ strided convolution in between them.  

The decoder starts with a transpose convolution, where the first decoder uses a stride value $4$, which increases the feature map spatial dimensions by $4$.  Similarly, the second decoder uses a stride value of $2$, increasing the spatial dimensions by $2$. Then, the network follows a simple concatenation and a residual block. Next, it is concatenated with the second skip connection and again followed by a residual block. The output of the last decoder block passes through a $1 \times 1$ convolution and a sigmoid activation function, generating the binary segmentation mask.



\subsubsection{Data Augmentation}
Supervised learning methods are data voracious and require large  amount of data to obtain reliable and well-performing models. Acquiring such training data through data collection, curation, and annotation is a manual process that needs significant resources and man-hours from both clinical experts and computational scientists. 

Data augmentation is a common technique to computationally increase the number of training samples in a dataset. For our \ac{DL} models, we use basic augmentation techniques such as horizontal flipping, vertical flipping, random rotation, random scale, and random cropping. The images used in all the experiments undergo normalization and are resized to a fixed size of $512\times512$. For the normalization, we subtract the image by mean and divide it by standard deviation.


\section{Results}
\label{sec:results}

In this section,  we first present our evaluation metrics and experimental setup. Then, we present both quantitative and qualitative results.
\subsection{Evaluation metrics}    

We have used standard computer vision metrics to evaluate polyp detection and localisation, and semantic segmentation methods on the Kvasir-SEG dataset. 

\subsubsection{Detection and localisation task}
For the object detection and localisation task, the commonly used Average Precision (AP) and \ac{IoU}  have been used~\cite{everingham2015pascal,lin2014microsoft}. 

\begin{itemize}
\item {IoU}: This metric measures the overlap between two bounding boxes A and B as the ratio between the overlapped area.
\begin{equation}
\text{IoU(A,B)} =\frac{A \cap B} {A \cup B} 
\end{equation}
\item {AP}: AP is computed as the Area Under Curve (AUC) of the precision-recall curve of detection sampled at all unique recall values (r1, r2, ...) whenever the maximum precision value drops:
\begin{equation}
    \mathrm{AP} = \sum_n{\left\{\left(r_{n+1}-r_{n}\right)p_{\mathrm{interp}}(r_{n+1})\right\}}, 
\end{equation}
with $p_{\mathrm{interp}}(r_{n+1}) =\underset{\tilde{r}\ge r_{n+1}}{\max}p(\tilde{r})$. Here, $p(r_n)$ denotes the precision value at a given recall value. This definition ensures monotonically decreasing precision. AP was computed as an average APs for IoU from 0.25 to 0.75 with a step-size of 0.05 which means an average over 11 IoU levels are used (AP @[.25 : .05 : .75]). 
\end{itemize}

\subsubsection{Segmentation task}
For polyp segmentation task, we have used widely accepted computer vision metrics that include \ac{DSC}, {Jaccard Coefficient} (JC), precision \textit{(p)}, and recall \textit{(r)}, and overall accuracy \textit{(Acc)}. JC is also termed as \ac{IoU}. We have also included \ac{FPS} to evaluate the clinical applicability of the segmentation methods in terms of inference time during the test. 

To define each metric, let \textit{tp}, \textit{fp}, \textit{tn}, and fn represents true positives, false positives, true negatives, and false negatives, respectively. 
\begin{equation}{\label{eq:1}}
\text{DSC} = \frac{2 \cdot tp} {2 \cdot tp + fp + fn}
\end{equation}
\begin{equation}
\text{IoU} = \frac{tp} {tp + fp + fn}
\end{equation}
\begin{equation}{\label{eq:1}}
\textit{r} = \frac{tp} {tp + fn}
\end{equation}

%
\begin{equation}{\label{eq:3}}
\textit{p} =\frac{tp} {tp + fp}
\end{equation}
\begin{equation}{\label{eq:1}}
\text{F2} = \frac{5p \times r} {4p + r}
\end{equation}
\begin{equation}{\label{eq:4}}
\textit{Acc} ={\frac{tp + tn} {tp + tn + fp + fn}} 
\end{equation}

\begin{equation}{\label{eq:7}}
\text{FPS} = {\frac{\#frames} {sec}} = {\frac{1} {sec /frame}}
\end{equation}

\begin{figure*} 
    \centering
    \includegraphics[width=\linewidth]{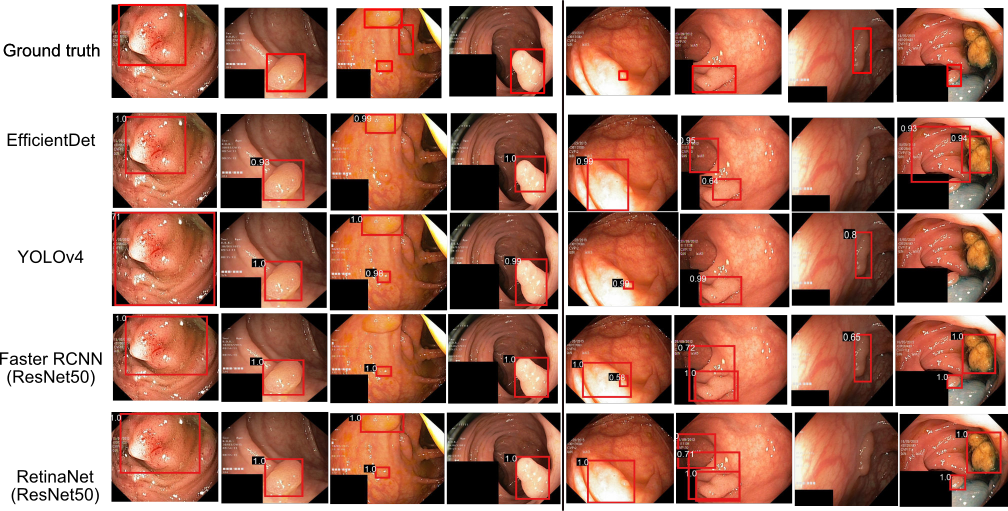}
    \caption{\textbf{Detection and localisation results on test dataset:} On right of the black solid line, images where EfficientDet-D0, YOLOv4, Faster R-CNN and RetinaNet (with ResNet50 backbone) have similar results and in most cases obtained highest IoU. On left, images with failed case (worse localisation) for either of the method. Confidence scores are provided on the top-left of the red prediction boxes.}
    \label{fig:detection}
\end{figure*}

\begin{table*}[t!]
\centering

\caption{{{Hyperparameters used for baseline methods for polyp detection and localisation task on Kvasir-SEG. Here,  CIoU: complete intersection-of-union loss, MSE: mean square error, CE: cross-entropy}}}   
\def\arraystretch{1.2}
\label{table:parametersuseddetection}
\begin{tabular}{c|c|c|c|c|c|c}
\toprule
\textbf{Method}  &\textbf{Learning rate} & \textbf{Optimizer} & \textbf{Batch size} & \textbf{Loss} & \textbf{Anchors} & \textbf{Threshold}\\ \midrule
Faster R-CNN~\cite{ren2015faster}    & 2.5e$^{-4}$  & Adam  &  8 &  L1$^{smooth}$, log-loss  & 256 &  0.4 \\ \hline
RetinaNet~\cite{lin2017focal}   & 1e$^{-5}$ &  SGD & 8  & L1$^{smooth}$, focal loss  &  15 (pyramid) & 0.3  \\ \hline
YOLOv3+spp~\cite{redmon2018yolov3}    &  1e$^{-3}$  & SGD  & 16  & MSE, 
CE  & 8 & 0.25  \\ \hline 
YOLOv4~\cite{bochkovskiy2020yolov4}   &   1e$^{-3}$ & SGD  & 16  & CIoU, CE    & 8 & 0.25  \\ \hline
EfficientDet-D0~\cite{tan2020efficientdet} & 1e$^{-4}$   & Adam  & 8  & Focal loss & default & 0.4  \\ 
\bottomrule
\end{tabular}
\end{table*}


\begin{table*}[t!]
\caption{Result on the polyp detection and localisation task on the Kvasir-SEG dataset. Two best scores are highlighted in bold.}
\label{tab:quantitative result}
\centering
\def\arraystretch{1.2}
\begin{tabular}{l|c|c|c|c|c|c|c}
\toprule
\textbf{Method} & \textbf{Backbone} & {\textbf{AP}}   & \textbf{IoU}  & \textbf{AP$_{25}$} & \textbf{AP$_{50}$} &\textbf{AP$_{75}$} &\textbf{FPS}\\ \midrule
EfficientDet-D0~\cite{tan2020efficientdet}
& EfficientNet-b0, biFPN & 0.4756 & 0.4322 &0.6846 &0.5047 & 0.2280 & 35.00\\ \hline

Faster R-CNN~\cite{ren2015faster} & ResNet50 & 0.7866 & 0.5621 & 0.8947 & 0.8418 & 0.5660 & 8.00\\ \hline

RetinaNet~\cite{lin2017focal} & ResNet50 &\textbf{0.8697} & 0.7313 & \textbf{0.9395} & \textbf{0.9095} & 0.6967 & 16.20\\ \hline
RetinaNet~\cite{lin2017focal} & ResNet101 & \textbf{0.8745} & 0.7579 & \textbf{0.9483} & \textbf{0.9095} &0.7132 & 16.80\\  \hline
YOLOv3+spp~\cite{redmon2018yolov3} & Darknet53 & 0.8105 & \textbf{0.8248} &0.8856 &\textbf{0.8532} & {\textbf{0.7586}} & 45.01\\
\hline
YOLOv4~\cite{bochkovskiy2020yolov4} & Darknet53, CSP & 0.8513 & 0.8025 &0.9123 & 0.8234 & \textbf{0.7594} & \textbf{48.00}\\
\hline
\textbf{ColonSegNet (Proposed)} & - & 0.8000 & \textbf{0.8100} & 0.9000 & 0.8166 & 0.6706 & \textbf{180.00}\\
\bottomrule
\end{tabular}
\end{table*}

\begin{table*}[t!h!]
\centering
\caption{Hyperparameters used for baseline methods for polyp segmentation task on Kvasir-SEG dataset}   
\def\arraystretch{1.2}
\label{table:parametersusedsegmentation}
\begin{tabular}{l|c|c|c|c|c|c|c}
\toprule
\textbf{Method} & \shortstack{\textbf{No. of }\\\textbf{parameters}} &\shortstack{\textbf{Learning}\\ \textbf{rate}} & \textbf{Optimizer} & \shortstack{\textbf{Batch}\\ \textbf{size}} & \textbf{Loss} & \textbf{Momentum} & \shortstack{\textbf{Decay}\\ \textbf{rate}}\\ \midrule

{UNet}~\cite{ronneberger2015u} & 7,858,433   &  1e${^{-2}}$ &  SGD & 8 & Cross-entropy  & - & -\\ \hline 

ResUNet~\cite{zhang2018road} &8,420,077 &1e${^{-4}}$ &Adam &8   &Dice loss   & - &- \\ \hline 

{ResUNet++}~\cite{jha2019resunet++} & 16,242,785 & 1e${^{-4}}$ &  Adam & 8 & Dice loss &- &-\\  \hline

HRNet~\cite{wang2020deep} &9,524,036   &1e${^{-4}}$ &Adam &8 &Dice loss & -&-\\  \hline 

{DoubleUNet}~\cite{jha2020doubleu} &  29,303,426 &   1e${^{-4}}$  &   Adam &  8 &  Dice loss &-  &- \\ \hline 
{PSPNet}~\cite{zhao2017pyramid} & 48,631,850   & 1e${^{-2}}$  &   SGD & 8 & Cross-entropy &-  & -\\  \hline  

{DeepLabv3+}~\cite{chen2018encoder}  & ResNet50: 39,756,962 & 1e${^{-2}}$  &   SGD & 8 & Cross-entropy & 0.9 & 1e${^{-4}}$  \\ \hline
  
{DeepLabv3+}~\cite{chen2018encoder} & ResNet101: 58,749,090 & 1e${^{-3}}$  &   SGD & 8 & Cross-entropy & 0.9 & 1e${^{-4}}$  \\  \hline

{FCN8}~\cite{long2015fully}  & 134,270,278  & 1e${^{-2}}$   & SGD  &  8  & Cross-entropy & 0.9 & 1e${^{-4}}$ \\  \hline
UNet-ResNet34 & 33,509,098   & 1e${^{-5}}$  & Adam  & 8  & Cross-entropy  &0.9  &1e${^{-4}}$  
 \\ \hline
 
ColonSegNet (Proposed) &5,014,049  &1e-4  &Adam &8  &Cross-entropy + Dice loss  &- &-  \\

 \bottomrule
\end{tabular}
\end{table*}

\begin{table*}[t!]
\centering
\def\arraystretch{1.2}
\caption{Baseline methods for polyp segmentation on the Kvasir-SEG dataset. Two best scores are highlighted in bold. "-" shows that there is no backbone used in the network. }
\label{res:polypsegmentation}
\begin{tabular}{l|l|c|c|c|c|c|c|c}
\toprule
\textbf{Method} & \multicolumn{1}{c|}{\textbf{Backbone}} & \textbf{Jaccard C.} & \textbf{DSC} & \textbf{F2-score} & \textbf{Precision} & \textbf{Recall} & \textbf{Overall Acc.} & \textbf{FPS} \\ \midrule
{UNet}~\cite{ronneberger2015u} & -  & 0.4713  & 0.5969    &0.5980   &0.6722  & 0.6171  & 0.8936 & 11.01\\ \hline

ResUNet~\cite{zhang2018road} & - & 0.5721 & 0.6902 & 0.6986  & 0.7454  & 0.7248  &0.9169 & 14.82 \\ \hline

{ResUNet++}~\cite{jha2019resunet++} & -  &   0.6126 &  0.7143  &  0.7198   &   0.7836     &   0.7419   & 0.9172  & 7.01 \\ \hline
{FCN8}~\cite{long2015fully}       &    VGG 16                                    & 0.7365      & 0.8310       & 0.8248      & 0.8817       & 0.8346       & 0.9524   &  24.91   \\ \hline
HRNet~\cite{wang2020deep} & -   &0.7592    &0.8446    &0.8467     & 0.8778     &0.8588 &0.9524 &11.69  \\ \hline

{DoubleUNet}~\cite{jha2020doubleu} & VGG 19   &   0.7332  &  0.8129  & 0.8207 &  0.8611   & 0.8402  & 0.9489  &  7.46   \\ \hline

{PSPNet}~\cite{zhao2017pyramid}     &   ResNet50                                     & 0.7444      & 0.8406       & 0.8314      & 0.8901       & 0.8357       & 0.9525    &  16.80  \\ \hline
{DeepLabv3+}~\cite{chen2018encoder} & ResNet50 &0.7759  &0.8572       & 0.8545 & 0.8907       & \textbf{0.8616}       & \textbf{0.9614} & 27.90    \\ \hline
{DeepLabv3+}~\cite{chen2018encoder} & ResNet101  & \textbf{0.7862}      & \textbf{0.8643}   & \textbf{0.8570}      & \textbf{0.9064}       & 0.8592       & 0.9608   &  16.75      \\ \hline

UNet~\cite{ronneberger2015u} & ResNet34 &\textbf{0.8100}  &\textbf{0.8757} &\textbf{0.8622 } & \textbf{0.9435} & \textbf{0.8597} & \textbf{0.9681 }&\textbf{35.00} \\ \hline

\textbf{ColonSegNet (Proposed)} &-  &0.7239  &0.8206 &0.8206  &0.8435  &0.8496 &0.9493  &{\textbf{182.38}} \\

\bottomrule 
\end{tabular}
\end{table*}

\begin{figure*}
    \centering
        \includegraphics[width=\linewidth]{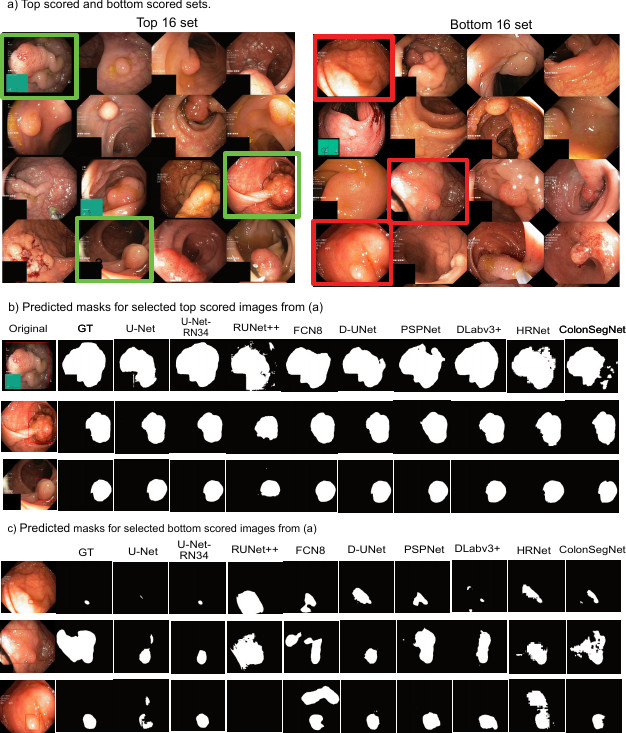}
    \caption{\textbf{Best and worse performing samples for polyp segmentation}: a) Top (left) and bottom (right) scored sets, b) predicted masks for top scored images and c) bottom scored images for all methods compared to the ground truth (GT) masks. Green rectangles represent the selected images from top scored set and red rectangle represent those from bottom set. Here, UNet-RN34: UNet-ResNet34, RUNet++: ResUNet++, D-UNet: Double UNet, DLabv3+: DeepLabv3+ (ResNet50).}
    \label{fig:bestandworst}
\end{figure*}
\subsection{Experimental setup and configuration}
The methods such as UNet, ResUNet, ResUNet++, DoubleUNet, and HRNet were implemented using Keras~\cite{chollet2015keras} with a Tensorflow~\cite{abadi2016tensorflow} back-end and were run on a Volta 100 GPU and an Nvidia DGX-2 AI system. A PyTorch  implementation for FCN8, PSPNet, DeepLabv3+, UNet-ResNet34, and ColonSegNet networks were done. Training of these methods were conducted on NVIDIA Quadro RTX 6000. NVIDIA GTX2080Ti was used for test inference for all methods reported in the paper. All of the detection methods were implemented using PyTorch  and used  NVIDIA Quadro RTX 6000 hardware for training the network. 

In all of the cases, we used 880 images for training and the remaining 120 images for the validation. Due to different image sizes in the dataset, we resized the images to $512\times 512$. Hyperparamters are important for the \ac{DL} algorithms to find the optimal solution. However, picking the optimal hyperparameter is difficult.  There are algorithms such as grid search, random search, and advanced solutions such as Bayesian optimization for finding the optimal parameters. However, an algorithm such as Bayesian optimization is computationally costly, making it difficult to test several \ac{DL} algorithms.  We have done an extensive hyperparameter search for finding the optimal hyperparameters for polyp detection, localisation, and segmentation task. These sets of hyperparameters were chosen based on empirical evaluation. The used hyperparameters are for the Kvasir-SEG dataset and are reported in the Table~\ref{table:parametersuseddetection}, and Table \ref{table:parametersusedsegmentation}. 

\subsection{Quantitative evaluation} 
\subsubsection{Detection and localisation} 
Table~\ref{tab:quantitative result} shows the detailed result for the polyp detection and localisation task on the Kvasir-SEG dataset. It can be observed that RetinaNet shows improvement over YOLOv3 and {YOLOv4} for mean average precision computed for multiple IoU thresholds and for average precision at \ac{IoU} threshold 25 (AP$_{25}$) and 50 (AP$_{50}$). RetinaNet with ResNet101 backbone achieved an average precision of 0.8745, while YOLOv4 yielded 0.8513. However, for the \ac{IoU} threshold of 0.75, YOLOv4 showed improvement over RetinaNet with (AP$_{75}$) of 0.7594 against 0.7132 for RetinaNet with ResNet101 backbone. Similarly, the average \ac{IoU} of 0.8248 was observed for YOLOv3, which is nearly 8\% improvement over RetinaNet. IoU determines the preciseness of the bounding box localisation. EfficientDet-D0 obtained the least AP of 0.4756 and \ac{IoU} of 0.4322. Faster R-CNN obtained an AP of 0.7866. However, it only obtained an \ac{FPS} of 8. YOLOv4 with Darknet53 as backbone obtained a \ac{FPS} of 48, which is 6$\times$ faster than Faster R-CNN. The other competitive network was YOLOv3, with an average \ac{FPS} of 45.01. However, its average precision value is 5\% less than YOLOv4. Thus, the quantitative results show that the YOLOv4 with Darknet can detect different types of polyps at a real-time speed of 48 \ac{FPS} and average precision of 0.8513. Therefore, from the evaluation metrics comparison, YOLOv4 with Darknet53 is the best model for detection and localisation of polyp.  The results suggest that the model can help gastroenterologists find missed polyps and decrease the polyp miss-rate. Even though, the proposed ColonSegNet is primary built for real-time segmentation of polyps, we compared the bounding box predictions of the proposed network with SOTA detection methods. It can be observed that  the inference of the proposed method is nearly four times faster (180 FPS) than YOLOv4. Additionally,  it is also obtaining competitive scores on both AP and IoU metrics (IoU of 0.81 and AP of 0.80). Therefore, it can also be considered as one of the best detection and localisation techniques.

\subsubsection{Segmentation} 
Table~\ref{res:polypsegmentation} shows the obtained results on the polyp segmentation task. It can be observed that the UNet with ResNet34 backbone performs better than the other SOTA segmentation methods in terms of \ac{DSC}, and \ac{IoU}. However, the proposed ColonSegNet outperforms in terms of processing speed. ColonSegNet is faster than UNet-ResNet34 by more than four times in processing colonoscopy frames. The complexity of the network is six times smaller than the UNet-ResNet34 network. The proposed network is even smaller than the conventional UNet, with its size only being around 0.75 times that of the UNet with higher scores on evaluation metrics compared to the classical UNet and its derivates such as ResUNet and ResUNet++. Additionally, the recall and overall accuracy metrics of ColonSegNet are close to the highest performing UNet-ResNet34 network, which shows the proposed method's efficiency.

The original implementation of UNet obtained the least \ac{DSC} of 0.5969, whereas the UNet with ResNet34 as the backbone model obtained the highest \ac{DSC} of 0.8757. The second and third best \ac{DSC} scores of 0.8643 and 0.8572 were obtained for DeepLabv3+ with ResNet101 and DeepLabv3+ with ResNet50 as the backbone, respectively. From the table, it is seen that DeepLabv3+ with ResNet101 performs better than Deeplabv3+ with ResNet50. This may be because of the top-5 accuracy (i.e., the validation results on the ImageNet model) of ResNet101 is slightly better than ResNet50\footnote{\url{https://keras.io/api/applications/}}. Despite of DeepLabv3+ with ResNet101 backbone having the total number of trainable parameters more than 11 times and DeepLabv3+ with ResNet34 being nearly eight times computational complexity, the DSC of ColonSegNet is competitive compared to both of these networks. However in terms, of processing speed, it is almost 11 times faster than DeepLabv3+ with ResNet101 and nearly seven times faster than DeepLabv3 with ResNet34 backbone. 

FCN8, HRNet and DoubleUNet provided similar results with DSC of 0.8310, 0.8446, and 0.8129 while ResUNet++ achieved DSC of only 0.7143. A similar trend can be observed for F2-score for all methods. For precision, UNet with ResNet34 backbone achieved the maximum score of $p =  0.9435$, and DeepLabv3+ with ResNet50 backbone achieved the highest scores of $r = 0.8616$, while UNet scored the worst with $p = 0.6722$ and $r = 0.6171$. The overall accuracy was outstanding for most methods, with the highest for UNet and ResNet34 as the backbone. IoU is also provided in the table for each segmentation method for scientific completion.  Again, UNet and ResNet34 surpassed others with a mIoU score of 0.8100. Also, UNet and ResNet34 achieved the highest \ac{FPS} rate of 35 fps, which is acceptable in terms of speed and is relatively faster as compared to  DeepLabv3+ with ResNet50 (27.9000) and DeepLabv3+ with ResNet101 (16.7500) and other SOTA methods. Additionally, when we consider the number of parameter uses (see Table~\ref{table:parametersusedsegmentation}), UNet with ResNet34 backbone uses less number of the parameters as compared to that of FCN8 or DeepLabv3+ network. Due to the low number of trainable parameters and fastest inference time, ColonSegNet  is computationally efficient and becomes the best choice while considering the need for real-time segmentation (182.38 FPS on NVIDIA GTX2080Ti) of polyps with deployment possible on even low-end hardware devices making it feasible for many clinical settings. Whereas, UNet with ResNet34 backbone seems the best choice while taking DSC metric into account, however, with speed of only 35 FPS on NVIDIA GTX2080Ti. 

%


\subsection{Qualitative evaluation} 
Figure~\ref{fig:detection} shows the qualitative result for the polyp detection and localisation task along with their corresponding confidence scores. It can be observed that for most images on the left side of the vertical line, both YOLOv4 and RetinaNet are able to detect and localise polyps with higher confidence, except for the third column sample where most of these methods can identify only some polyp areas. Similarly, on the right side of the vertical line, the detected bounding boxes for 5th and 6th column images are too wide for the RetinaNet, while YOLOv4 has the best localisation of polyp (observe the bounding box). Also, in the seventh column, RetinaNet and EfficientDet D0 misses the polyp. In the eighth column, YOLOv4 and EfficientDet D0 misses the small polyp completely while stool and polyp is detected as polyp by the Faster R-CNN and RetinaNet. Figure~\ref{fig:bestandworst} shows the result for the top-scored and bottom scored sets selected based on their dice similarity coefficient values for the semantic segmentation methods. It can be seen that all the algorithms are able to detect large polyps and produce high-quality masks (see Figure~\ref{fig:bestandworst}(b).

Here, the best obtained segmentation results can be observed for DeepLabv3+ and UNet-ResNet34. However, as shown in Figure~\ref{fig:bestandworst}(c), the segmentation results are affected for flat polyps (very small), images with a certain degree of inclined view, and for the images with saturated areas. The proposed ColonSegNet is able to achieve similar shapes compared to these of the ground truth with some outliers for the predictions which can be seen in Figure~\ref{fig:bestandworst}(b), while for the prediction on worse performing images in Figure~\ref{fig:bestandworst}(c), our proposed network provides comparatively improved predictions on almost all samples.

\section{Discussion}
\label{sec:discussion}
It is evident that there is a growing interest in the investigation of computational support systems for decision making through endoscopic images. For the first time, we are using Kvasir-SEG for detection and localisation tasks, and comparing segmentation methods with most recent SOTA methods. We provide a reproducible benchmarking of the \ac{DL} methods using standard computer vision metrics in object detection and localisation, and semantic segmentation. The choice of methods are based their popularity in the medical image domain for detection and segmentation (e.g., UNet, Faster R-CNN), speed (e.g., UNet with ResNet34, YOLOv3), and accuracy (e.g., PSPNet, FCN8, or DoubleUNet) or a combination of all (e.g., DeepLabv3+, YOLOv4). 

From the experimental results in Table~\ref{tab:quantitative result}, we can observe that the combination of YOLOv3 with Darknet53 backbone shows improvement over other methods in terms of mIoU, which means a better localisation compared to counterpart RetinaNet. However, YOLOv4 is 3$\times$ faster than RetinaNet and has a good trade-off between the average precision and IoU. This is because of their Cross-Stage-Partial-Connections (CSP) and CIoU loss for bounding box regression. However, RetinaNet with the backbone ResNet101 shows competitive results surpassing other methods on average precision but nearly 5\% less IoU compared to YOLOv4 and nearly 5\% less than YOLOv3-spp. Similarly, state-of-the-art methods Faster R-CNN and EffecientDet-D0 provided the least AP and IoU.

A choice between computational speed, accuracy and precision is vital in object detection and localisation tasks, especially for colonoscopy video data where speed is a vital element to achieve real-time performance. Therefore, we consider YOLOv4 with Darknet53 and CSP backbone as the best approach in the table for the polyp detection and localisation task.

For the semantic segmentation tasks, ColonSegNet showed improvement over all the methods. The method obtained the highest FPS of 182.38. The quantitative results in Figure~\ref{fig:bestandworst} (b) showed the most accurate delineation of polyp pixels compared to other SOTA methods considered in this paper. The most competitive method to ColonSegNet was UNet with ResNet34 backbone. The other comparable method was DeepLabv3+, which accuracy can be due to its ability to navigate the semantically meaningful regions with its atrous convolution and spatial-pyramid pooling mechanism. Additionally, the feature concatenation from previous feature maps may have helped to compute more accurate maps for object semantic representation and hence segmentation. The other competitor was PSPNet, which is also based on similar idea but on aggregating the global context information from different regions rather than the use of dilated convolutions. The computational speed for DeepLabv3+ with the same ResNet50 backbone as used in PSPNet in our experiments comes from the fact that the 1D separable convolutions and SPP network is used in DeepLabv3+. We evaluated the most recent popular SOTA method in segmentation ``HRNet''~\cite{wang2020deep}. While HRNet produced competitive results compared to other SOTA methods, UNet with ResNet34 backbone and DeepLabv3+ outperformed for most evaluation metrics with ColonSegNet being competitive in the recall, and overall accuracy and outperforming other SOTA method significantly. 

Figure~\ref{fig:bestandworst} shows an example for the 16 top scored and 16 bottom scored images on \ac{DSC} for segmentation. From the results in Figure~\ref{fig:bestandworst}(c), it can be observed  that there are polyps whose appearance under the given lighting conditions is very similar to healthy surrounding gastrointestinal skin texture. We suggest that including more samples with variable texture, different lighting conditions, and different angular views (refer to the samples in Figure~\ref{fig:bestandworst}(a) on the right, and (c)) can help to improve the \ac{DSC} and other metrics of segmentation. We also observed  that the presence of sessile or flat polyps were major limiting factors for algorithm robustness. Thus, including smaller polyps with respect to image size can help algorithm to generalise better thereby making these methods more usable for early detection of hard-to find polyps. In this regard, we also suggest the use of spatial pyramid layers to handle small polyps and using context-aware methods such as incorporation of artifacts or shape information to improve the robustness of these methods.

The possible limitation of the study is its retrospective design. Clinical studies are required for the validation of the approach in a real-world setting~\cite{mori2018real}. Additionally, in the presented study design we have resized the images, which can lead to loss of information and affect the algorithm performance. Moreover, we have optimized all the algorithms based on the empirical evaluation. Even though, optimal hyper-parameters have been set after experiments, we acknowledge that these can be further adjusted. Similarly, meta-learning approaches can be exploited to optimize the hyper-parameters that can work even in resource constraint settings.
\section{Conclusion}
\label{sec:conclusion}
 In this paper, we benchmark deep learning methods on the Kvasir-SEG dataset. We conducted thorough and extensive experiments for polyp detection, localisation, and segmentation tasks and shown how different algorithms performs on variable polyp sizes and image resolutions. The proposed ColonSegNet detected and localised polyps at 180 frames per second. Similarly, ColonSegNet segmented polyps at the speed of 182.38 frames per second. The automatic polyp detection, localisation, and segmentation algorithms showed good performance, as evidenced by high average precision, IoU, and \ac{FPS} for the detection algorithm and DSC, IoU, precision, recall, F2-score, and FPS for the segmentation algorithm. While algorithms investigated in this paper show a clear strength to be used in clinical settings to help gastroenterologists for the polyp detection, localisation, and segmentation task, computational scientists can build upon these methods to further improve in terms of accuracy, speed and robustness. 

Additionally, the qualitative results provide insight for failure cases. This gives an opportunity to address the challenges present in the Kvasir-SEG dataset.  Moreover, we have provided experimental results using well-established performance metrics along with the dataset for a fair comparison of the approaches. We believe that further data augmentation, fine tuning, and more advanced methods can improve the results. Additionally, incorporating artifacts~\cite{ali2021deep} (e.g., saturation, specularity, bubbles, and contrast) issues can help improve the performance of polyp detection, localisation, and segmentation. In the future, research should be more focused on designing even better algorithms for detection, localisation, and segmentation tasks, and models should be build taking the number of parameters into consideration as required by most clinical systems.

\section*{Acknowledgement}
D. Jha is funded by  Research Council of Norway project number 263248 (Privaton). The computations in this paper were performed on equipment provided by the Experimental Infrastructure for Exploration of Exascale Computing (eX3), which is financially supported by the Research Council of Norway under contract 270053. Parts of computational resources were also used from the research supported by the National Institute for Health Research (NIHR) Oxford BRC with additional support from the Wellcome Trust Core Award Grant Number 203141/Z/16/Z. S. Ali is supported by the NIHR Oxford Biomedical Research Centre. The views expressed are those of the author(s) and not necessarily those of the NHS, the NIHR or the Department of Health.

\begin{IEEEbiography}[{\includegraphics[width=1in,height=1.25in,clip,keepaspectratio]{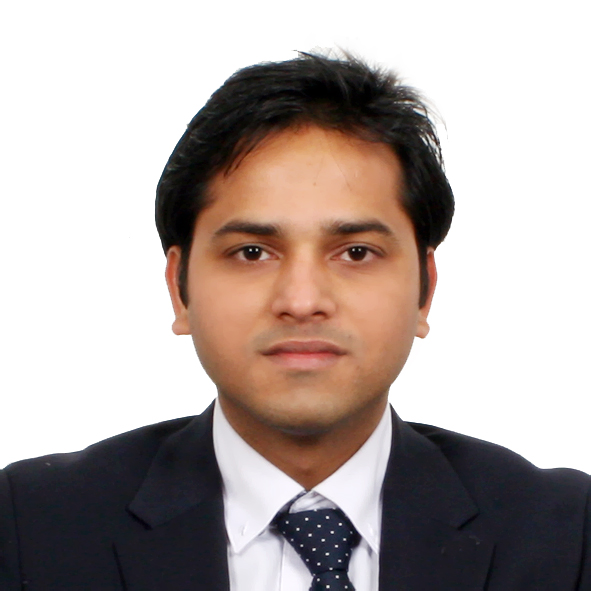}}]{Debesh Jha}  received his Master's Degree in Information and Communication Engineering from Chosun University, Gwangju, Republic of Korea. He is currently pursuing a Ph.D. with the SimulaMet, Oslo, Norway and UiT The Arctic University of Norway, Tr{\o}mso, Norway. His research interest includes computer vision, machine learning, deep learning, and medical image analysis.
\end{IEEEbiography}

\begin{IEEEbiography}[{\includegraphics[width=1in,height=1.25in,clip,keepaspectratio]{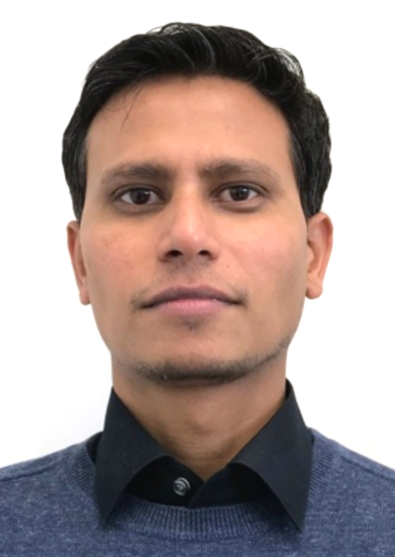}}]{Sharib Ali} received his Ph.D from University of Lorraine, France. He is currently working at the Institute of Biomedical Engineering, Department of Engineering Science, University of Oxford, Oxford, UK. Previously, he also worked as a post doctoral researcher at the Biomedical Computer Vision Group and German Cancer research Center (DKFZ), University of Heidelberg, Heidelberg, Germany. His research interests include computer vision and medical image analysis.
\end{IEEEbiography}

\begin{IEEEbiography}[{\includegraphics[width=1in,height=1.25in,clip,keepaspectratio]{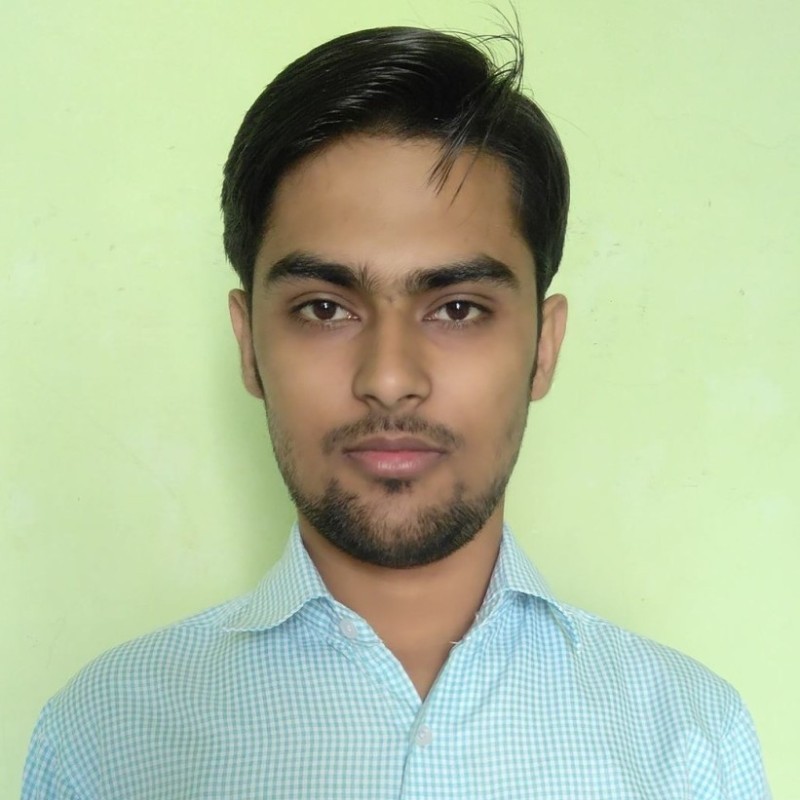}}]{Nikhil Kumar Tomar} received his Bachelor's Degree in Computer Application from Indira Gandhi Open University, Delhi, India. He is currently doing collaborative research with SimulaMet. His research interest includes computer vision,  artificial intelligence, parallel processing, and medical image segmentation.
\end{IEEEbiography}

\begin{IEEEbiography}[{\includegraphics[width=1in,height=1.25in,clip,keepaspectratio]{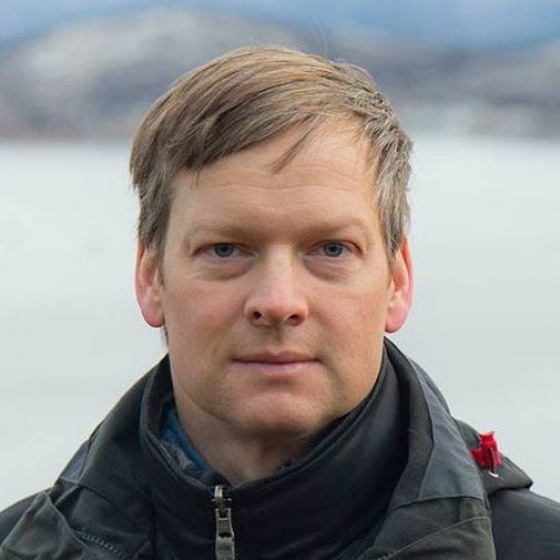}}]{H{\AA}vard D. Johansen} received his Ph.D from the UiT  The Arctic University of Norway. He is a professor at the Department of Informatics, UiT The Arctic University of Norway. His major research involves computing Network, Cloud Computing, Network Security, Information Security, Network architecture. 
\end{IEEEbiography}

\begin{IEEEbiography}[{\includegraphics[width=1in,height=1.25in,clip,keepaspectratio]{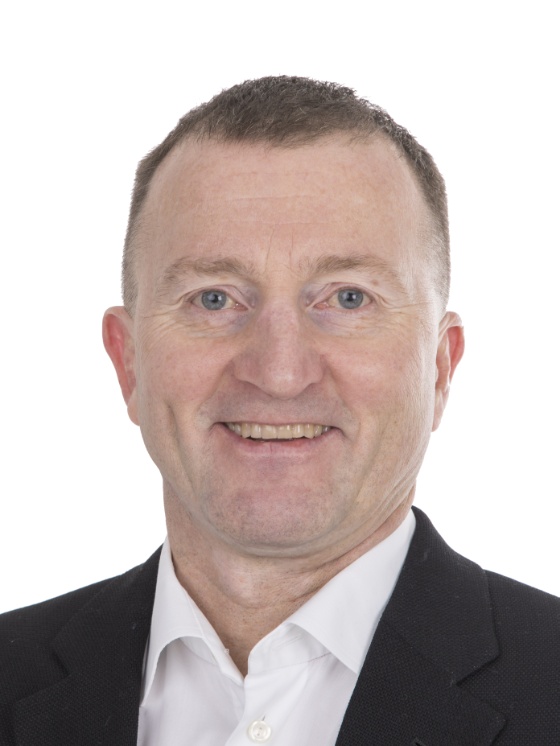}}]{Dag Johansen} is a Full professor at the department of Computer Science, UiT The Arctic of Norway. He is exploring interdisciplinary research problems at the intersection of sport science, medicine, and computer science. A use-case receiving special attention is elite soccer performance development and quantification technologies as basis for evidence-based decisions. His research focus is on intervention technologies where privacy is a first-order concern and design principle.
\end{IEEEbiography}

\begin{IEEEbiography}[{\includegraphics[width=1in,height=1.25in,clip,keepaspectratio]{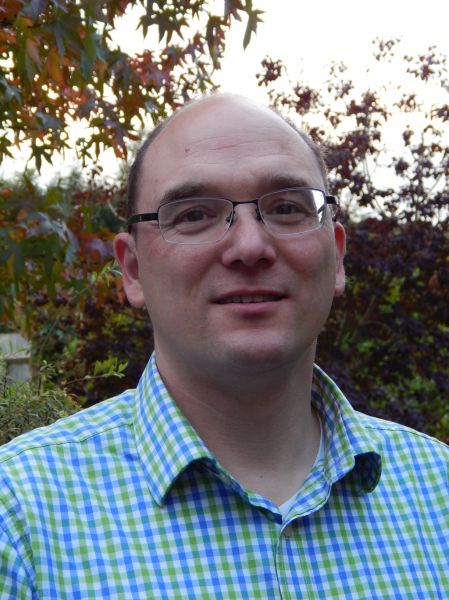}}] {Jens Rittscher} is professor at the Institute of Biomedical Engineering, Department of Engineering science, University of Oxford, Oxford, UK. Prof. Rittscher has worked extensively in the area of video surveillance, the automatic annotation of video, and understanding of volumetric seismic data. He previously worked at the GE Global Research in Niskayuna, NY, USA where he led the Computer Vision Laboratory. Prof. Rittscher obtained his PhD from the University of Oxford in 2001. He is a member of IEEE and acts as an elected member of the IEEE SPS Technical Committee on Bio Image and Signal Processing.
\end{IEEEbiography}

\begin{IEEEbiography}[{\includegraphics[width=1in,height=1.25in,clip,keepaspectratio]{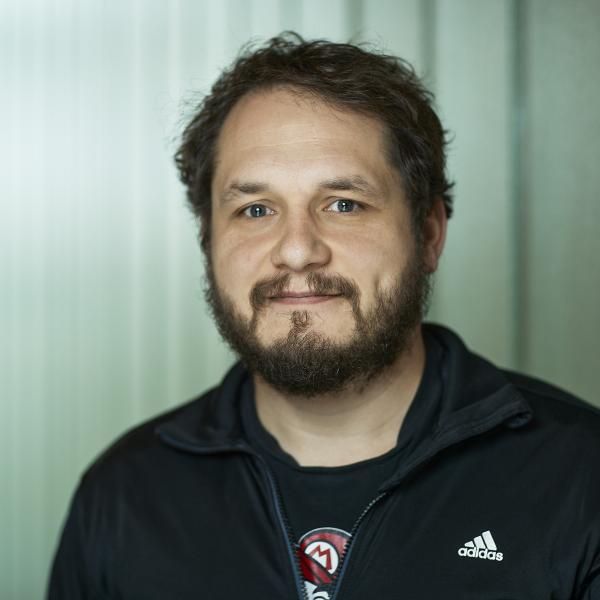}}]{Michael A. Riegler} received PhD degree in the department of informatics from University of Oslo, Oslo, Norway, in 2015. He is currently working as chief research scientist at SimulaMet, Oslo, Norway. His research interests include machine. learning, video analysis and understanding, image processing, image retrieval, crowdsourcing, social computing, and user intentions.
\end{IEEEbiography}

\begin{IEEEbiography}[{\includegraphics[width=1in,height=1.25in,clip,keepaspectratio]{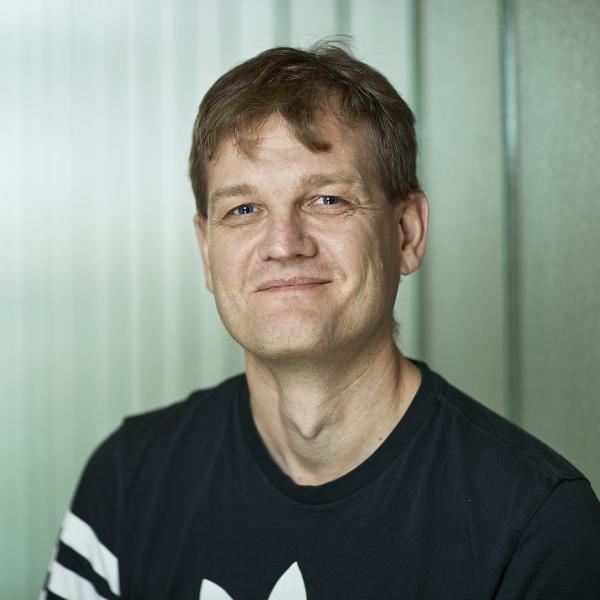}}]{P{\AA}l Halvorsen} is a chief research scientist at the SimulaMet, Oslo, Norway, and a full Professor with the Department of Computer Science, Oslo Metropolitan University, and an adjunct Professor at Department of Informatics, University of Oslo, Norway. His research interest includes distributed multimedia systems, including operating systems, processing, storage and retrieval, communication, and distribution.
\end{IEEEbiography}

\bibliographystyle{IEEEtran}
\bibliography{references} 
\end{document}